%% file: latex/main.tex
\documentclass[11pt]{article}

\usepackage[final]{acl}

\usepackage{amsfonts}
\usepackage{amssymb}

\usepackage{times}
\usepackage{latexsym}

\usepackage[T1]{fontenc}
\usepackage[utf8]{inputenc}

\usepackage{microtype}

\usepackage{inconsolata}

\usepackage{graphicx}
\usepackage{multirow}
\usepackage{booktabs}
\usepackage{amsmath} 
\usepackage{booktabs}
\usepackage[table]{xcolor}
\newcommand{\mydeltax}[1]{\cellcolor{gray!10}\textbf{#1}}
\newcommand{\mydelta}[1]{\cellcolor{gray!10}\textbf{#1}}

\title{\textsc{ProEvent}: An Event-centric Benchmark for Proactive Agents}

\author{
  Guanzhen Li \quad Liangming Pan \quad Leye Wang \\
  School of Computing, Peking University \\
  gzli25@stu.pku.edu.cn
}

\begin{document}
\maketitle
\raggedbottom
\input{Body/Abstract}

\input{Body/Introduction}

\input{Body/Related}

\input{Body/Problem}
\input{Body/ProEvent}

\input{Body/Experiment}

\input{Body/Conclusion}

\section*{Limitations}

While constructing \textsc{ProEvent}, we generate chats using LLMs. Although we promote diversity by varying event types and incorporating real-world dynamics such as negotiation complexity, concurrency, and noise, the generated dialogues still tend to be structured and repetitive, lacking the spontaneity and variability of human conversations. Moreover, real-world event planning often involves more complex coordination across multiple chats and participants, whereas we currently assume that all negotiations for an event occur within a single chat. In future work, we plan to refine and expand \textsc{ProEvent} to further enhance linguistic diversity and interaction complexity.

% \section*{Acknowledgments}

\bibliography{custom}

\appendix
\input{Appendix/Appendix_A}

\end{document}

%% file: Body/Abstract.tex
\begin{abstract}

Proactive agents are expected to anticipate user needs and provide autonomous assistance by perceiving environmental context without explicit instructions.
A fundamental capability of such agents is to identify and track users’ upcoming events, enabling continuous and event-specific assistance.
For example, by recording the time and location of a planned hike, an agent can deliver weather reminders in advance or provide navigation support before departure.
However, existing works on proactive agents largely overlook event-centric assistance, and the open-ended nature of proactive assistance poses challenges for reliable evaluation.

To bridge these gaps, we introduce \textsc{ProEvent}, the first event-centric benchmark designed to assess an agent’s ability to proactively maintain a user’s timetable based on ongoing instant messaging chats.
\textsc{ProEvent} provides synthesized yet realistic chats that consider the dynamic interaction among users, concurrent chat threads, and noise in the real world, and evaluates proactive agents on response timing, single-step response correctness, and multi-step response correctness.
Experiments on eight LLMs and pipelines reveal that current agents frequently overact and struggle with event cancellation.
Notably, even GPT-5.1 only reacts correctly in $26.7\%$ of scenarios.
Further qualitative analysis reveals fundamental limitations of current LLMs as proactive agents, particularly in detecting implicit events and reasoning from the user’s first--person perspective.

\end{abstract}

%% file: Body/Introduction.tex
\section{Introduction}

\begin{figure*}[]
  \includegraphics[width=\linewidth]{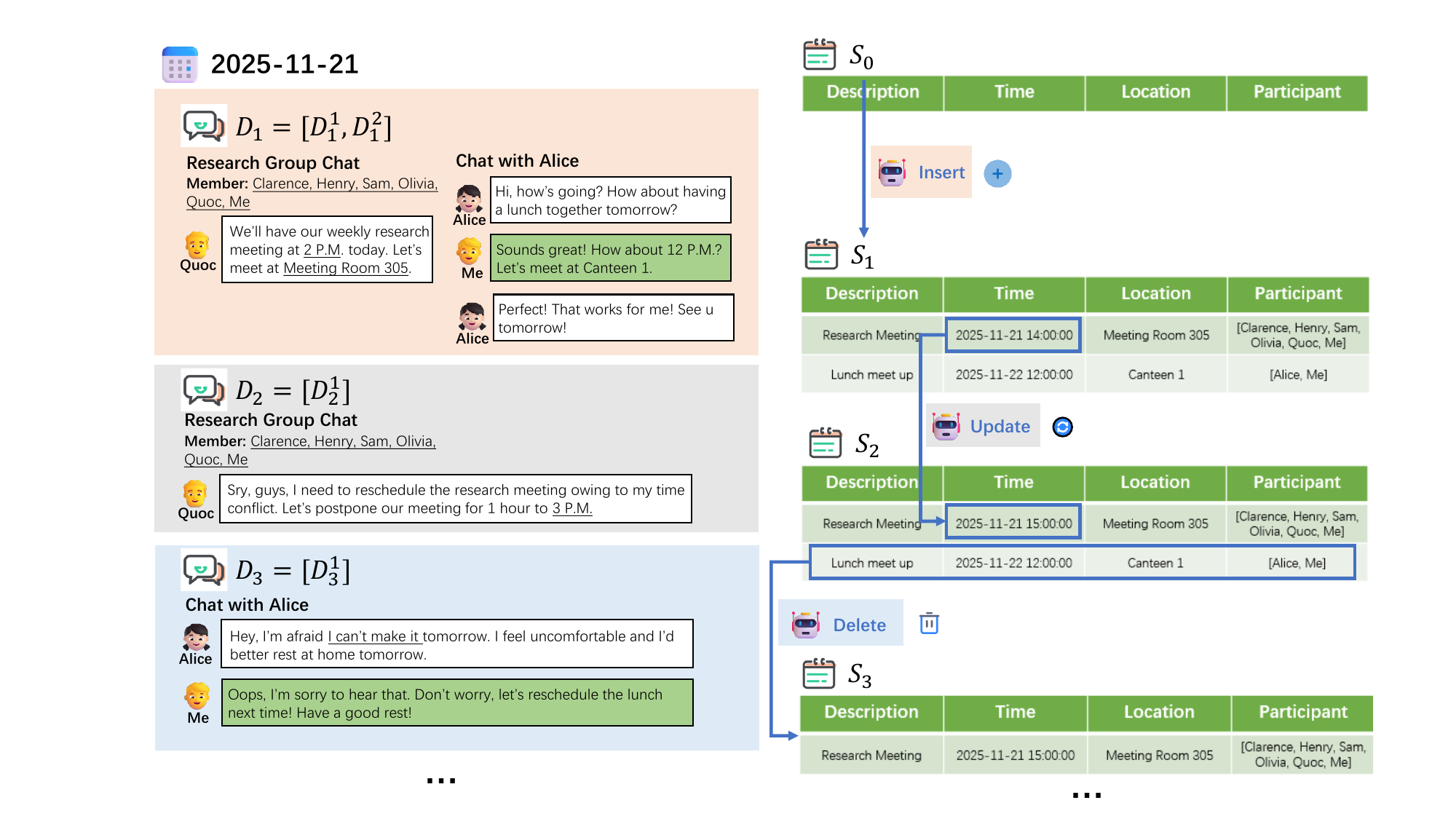}
  \caption{Illustration of the timetable maintenance task. The proactive agent needs to maintain the timetable on the right by proactively seeking information from the chats on the left. $D_t$ refers to the messages received between $t-1$ and $t$. $S_t$ refers to the user's timetable at time $t$.}
  \label{fig1}
\end{figure*}

% \textbf{Para 1:} Proactive Agent Background.
% Recent advances in large language models (LLMs) have enabled agentic systems to support humans in diverse tasks, such as planning~\cite{xie2024travelplanner,zhu2025knowagent}, task execution~\cite{zhang2025webpilot,sun2025genesis}, and more. However, most existing LLM-based agents remain reactive, largely relying on explicit user instructions, which may result in repetitive human intervention and the risk of missing time-sensitive information. To address these limitations, it is essential to develop proactive agents that can autonomously perceive and respond to their environment.
Recent advances in large language models (LLMs) have enabled agentic systems to support humans in diverse tasks~\cite{xie2024travelplanner,zhang2025webpilot}. However, most existing LLM-based agents remain \textit{reactive}, largely relying on explicit user instructions, which may result in repetitive human intervention and the risk of missing time-sensitive information. To address these limitations, it is essential to develop \textit{proactive} agents that can autonomously perceive and respond to their environment.

Events, defined as the activities a user plans to attend, are critical for proactive agents. For one thing, anticipating a user's upcoming events enables proactive agents to \textbf{provide continuous and seamless assistance} in advance. For example, when a user schedules a hiking trip, recording the time and location allows agents to proactively remind the user of weather conditions in advance, as well as offer navigation assistance before departure. For another thing, maintaining event records enables agents to deliver assistance \textbf{tailored to specific activities}. For instance, a proactive agent may suggest reserving a table for a planned meal or remind the user to prepare for an upcoming meeting.
% Events, which refer to anticipated user activities, are fundamental to proactive agents. By leveraging event information, a proactive agent can (1) anticipate upcoming activities and respond in advance, and (2) generate tailored, context-aware responses that are specific to the requirements of each event. For example, given a scheduled meeting, the agent can remind the user of the meeting time in advance and prompt the user to prepare accordingly.

There have been several proactive agent benchmarks focusing on human-–device interactions~\cite{lu2024proactive,yang2025fingertip,zhao2025proactiveva,pasternak2025beyond}, aiming to provide immediate assistance for users (e.g., summarizing a webpage when the user opens it). However, they suffer from two limitations. First, to the best of our knowledge, \textbf{no existing benchmarks evaluate proactive agents’ ability to track users’ upcoming events}, which limits their capacity to provide continuous and long-term assistance for future activities. Second, most existing benchmarks \textbf{lack reliable evaluation protocols} due to the open-ended nature of the predicted responses. They typically rely on semantic similarity~\cite{yang2025fingertip} or use LLMs to simulate a human judge~\cite{lu2024proactive,pasternak2025beyond}, which may not accurately reflect the correctness or usefulness of the agent response.

In light of these limitations, we propose an Event-centric Benchmark for Proactive Agents (\textsc{ProEvent}). \textsc{ProEvent} focuses on a concrete task: maintaining a user's timetable by proactively collecting event information from the ongoing instant messaging chats. We focus on instant messaging chats because they are ubiquitous in the real world yet highly challenging, due to \textbf{dynamic speaker interactions}, \textbf{concurrent chat threads}, and \textbf{pervasive noise}~\cite{zhang2020recent,sapkota2025multi}. To construct \textsc{ProEvent}, we develop a data generation pipeline that synthesizes realistic chats with rigorous human quality control. Furthermore, we provide a comprehensive evaluation suite that assesses both the timeliness and the correctness of proactive responses.
% \textbf{Para 4:} (1) We propose ProEvent, an event-centric benchmark for agents to proactively collect event information from the chat logs. (2) To construct ProEvent, we synthesize realistic and high-quality chat logs. (3) We develop comprehensive evaluation metrics for quantitative assessment of proactive agents.
% In light of the above limitations, we introduce \textsc{ProEvent}, an event-centric benchmark for proactive agents that focuses on a specific but common task--maintaining the user's timetable. 
% This task evaluates proactive agents' capabilities from various perspectives: (1) proactively deciding whether to respond and how to respond based on the newly received messages without explicit user instructions; (2) reasoning over noisy chat environments, including irrelevant messages, informal expressions, and long time intervals between related messages; and (3) anticipating users' future activities, thereby enabling timely and seamless assistance for certain events.
% To construct this benchmark, we synthesize realistic chat conversations and their corresponding evolving timetables. Unlike existing works based on human evaluation or LLM evaluation, \textsc{ProEvent} supports objective and fine-grained evaluation on proactive agents' response timing and response correctness by comparing the prediction and ground-truth timetables.

In this work, using \textsc{ProEvent}, we evaluate eight LLMs and pipelines, and find that they tend to overreact for users and struggle with event cancellation scenarios. Furthermore, we show that real-world chat complexities, including dynamic interactions, concurrent chats, and noise, all substantially challenge LLMs’ performance as proactive agents. Our qualitative analysis further reveals key deficiencies in current LLMs, including difficulties in capturing implicit events and making decisions from the user’s first-person perspective.

%% file: Body/Related.tex
\section{Related Work}

\paragraph{Proactive Agent Benchmark.}
Recently, several benchmarks have been proposed to evaluate proactive agents. ProactiveBench~\cite{lu2024proactive} and FingerTip~\cite{yang2025fingertip} formulate proactive response as a sequential prediction problem, predicting the user’s next-step action, whereas PROBE~\cite{pasternak2025beyond} provides rich contextual information and requires agents to address users’ potential needs with tools.
Beyond benchmarks, ProactiveVA~\cite{zhao2025proactiveva} and ProAgent~\cite{yang2025contextagent} propose proactive agent pipelines for GUI operations and smart glasses, evaluated in real-world settings with feedback from users.
All these works evaluate both response timing and response correctness, with correctness being more challenging to assess due to the open-ended nature of the response.
ProactiveBench~\cite{lu2024proactive} addresses this by fine-tuning a small model to simulate human judgment, whereas FingerTip~\cite{yang2025fingertip} relies on semantic similarity between predictions and ground truth. However, both approaches lack objective grounding for evaluation. 
To address this limitation, \textsc{ProEvent} introduces a comprehensive evaluation suite that enables objective assessment of response correctness based on the correctness of the user's timetable.
In addition, with respect to data sources, most of existing works face challenges in acquiring real--world data and therefore rely on synthetic datasets. While we also synthesize chats, we design a pipeline to maximize chat realism and ensure high data quality through rigorous quality control.

\paragraph{Chat Synthesis with LLMs}
Prior works have applied LLMs to synthesize realistic dialogues by firstly defining the role profiles and then guiding the interactions between roles with structured prompts~\cite{wu2023large,qiu2024interactive,wang2023voyager,park2023generative}.
% LLMs have been widely used for role-playing~\cite{shanahan2023role,shao2023character} and dialogue synthesis~\cite{abdullin2023synthetic,wang2025toolflow}. Prior work has explored generating realistic dialogues in various domains, such as healthcare~\cite{wu2023large,qiu2024interactive} and games~\cite{wang2023voyager,park2023generative,wu2024role}, by defining role profiles and guiding interactions with structured prompts. 
% In addition, LLMs have also been used to generate dialogues conditioned on external knowledge~\cite{suresh2025diasynth} or explicit intentions~\cite{fu2023improving}.
While these works provide useful insights, synthesizing realistic instant messaging chats must capture additional key characteristics, including interaction dynamics, concurrent chat threads, and pervasive noise~\cite{zhang2020recent,sapkota2025multi}.
Specifically, in negotiation scenarios, conversational goals often evolve over time, resulting in dynamic interactions among speakers that are challenging for models to accurately track the current interaction state~\cite{peng2018deep,wu2019switch}.
Moreover, prior work has shown that handling concurrent chats introduces additional complexity~\cite{sapkota2025multi}, while the noise can obscure critical information and distract models~\cite{yang2023dialogue,sapkota2025multi}.
% Besides, real-world chats contain substantial noise, including off-topic messages, informal language, and more~\cite{yang2023dialogue,sapkota2025multi}.
% While such phenomena are common in chat logs, they are rarely considered in conventional dialogue generation.
Accordingly, to account for these challenges, we synthesize realistic chats with multiple negotiation turns during event planning, construct scenarios with varying numbers of concurrent chats, and inject diverse noise, as detailed in Section~\ref{proevent}.
% we first define a set of contact profiles and guide chat generation with predefined intentions for planning specific events. To better reflect the characteristics of real-world chats, we generate conversations with varying numbers of turns, simulate scenarios involving concurrent conversations, and inject diverse forms of noise. These designs will be detailed in Section~\ref{proevent}.

%% file: Body/Problem.tex
\section{Problem Formulation and Evaluation}

\subsection{Problem Formulation}

As \textsc{ProEvent} introduces a novel task, we first formalize the problem and then present the corresponding evaluation suite in next section.

As shown in Figure~\ref{fig1}, given a set of $N$ chats $D_t=\{D_t^1, D_t^2, \ldots, D_t^N\}$ received by the user between time $t-1$ and time $t$, the goal of \textsc{ProEvent} is to update the user's timetable from the previous state $S_{t-1}$ to a new state $S_t$.

\begin{equation}
S_t = f(S_{t-1}, D_t)
\end{equation}

Each timetable $S_t=\{E_1, E_2, \ldots, E_M\}$ consists of $M$ events. Following the widely adopted iCalendar data format, each event contains structured attributes including start time, end time, location, participants, and a brief description. Each chat $D_t^i$ is a multi-turn dialogue between the user and a contact or within a group conversation. 

To better reflect real-world proactive agent settings (e.g., GPT-Pulse\footnote{\url{https://openai.com/zh-Hans-CN/index/introducing-chatgpt-pulse/}} and Mine Context\footnote{\url{https://github.com/volcengine/MineContext}}), we adopt a discrete-time formulation in which agents collect environmental context changes within fixed time intervals. Hence, all chats received within the same interval are processed jointly, and a single event may be updated across multiple time steps.

To avoid repeatedly outputting unchanged events at each time step, we require agents to generate explicit \emph{timetable operations} rather than directly outputting the updated timetable. We define three operations: \textit{Insert}, \textit{Update}, and \textit{Delete}.

\textbf{Insert}(\textit{start\_time}, \textit{end\_time}, \textit{location}, \textit{participants}, \textit{description}): Insert a new event into the timetable with the specified attributes.

\textbf{Update}$(\textit{id}, \textit{attribute}, \textit{value})$: Modify the specified attribute of the event identified by $\textit{id}$ using the provided value.

\textbf{Delete}$(\textit{id})$: Remove the event with the specified $\textit{id}$ from the timetable.

Hence, at each time step $t$, the agent outputs a list of timetable operations.

\subsection{Evaluation Suite}

Given an agent’s predicted operations at time $t_i$, denoted as $P_{t_i}$, and the ground-truth operations $G_{t_i}$, we evaluate proactive agents from three perspectives:
(1) response timing, (2) single-step correctness, and (3) multi-step correctness.

\paragraph{Response Timing.}
Response timing evaluates whether the agent triggers assistance or stays await at appropriate time steps. We adopt the following two metrics.

\emph{False Detection Rate (FDR)} measures the proportion of time steps where the agent produces operations when no ground-truth operation is required:
\begin{equation}
\mathrm{FDR} =
\frac{
\sum_{t_i} \mathbb{I}\big(|P_{t_i}| > 0 \land |G_{t_i}| = 0\big)
}{
\sum_{t_i} \mathbb{I}\big(|G_{t_i}| = 0\big)
}.
\end{equation}
\emph{Missed Need Rate (MNR)} measures the proportion of time steps where the agent fails to respond when ground-truth operations exist:
\begin{equation}
\mathrm{MNR} =
\frac{
\sum_{t_i} \mathbb{I}\big(|P_{t_i}| = 0 \land |G_{t_i}| > 0\big)
}{
\sum_{t_i} \mathbb{I}\big(|G_{t_i}| > 0\big)
}.
\end{equation}

\paragraph{Single--step Response Correctness.}
At each time step $t_i$, the agent predicts a set of operations $P_{t_i}$, which is compared against the ground truth set $G_{t_i}$. To evaluate the response correctness at each time step, we adopt \emph{Precision} and \emph{Recall}:
\begin{equation}
\mathrm{Precision} =
\frac{
\sum_{t_i} |P_{t_i} \cap G_{t_i}|
}{
\sum_{t_i} |P_{t_i}|
},
\end{equation}
\begin{equation}
\mathrm{Recall} =
\frac{
\sum_{t_i} |P_{t_i} \cap G_{t_i}|
}{
\sum_{t_i} |G_{t_i}|
}.
\end{equation}

\paragraph{Multi--step Response Correctness.}

We evaluate response correctness in a multi step setting using two metrics: \emph{Event Success Rate (ESR)} and \emph{Timetable Success Rate (TSR)}. \emph{ESR} measures the proportion of events that are correctly predicted after multiple steps of accumulation, where an event is considered correct only if all its attributes match the ground truth. \emph{TSR} measures the proportion of completely correct timetables, where a timetable is considered correct only when all events it contains are correct. 
Notably, beyond evaluating multi--step response correctness, \emph{ESR} and \emph{TSR} are also designed to resolve ambiguities in single--step evaluation. For example, an \emph{Update} operation may lead to the same final event state as a \emph{Delete} followed by an \emph{Insert}. Although the two responses differ at the single-step level, both should be considered correct. Evaluation on success rate resolves this ambiguity.

%% file: Body/ProEvent.tex
\section{\textsc{ProEvent}}
\label{proevent}

\begin{figure*}[t]
  \includegraphics[width=\linewidth]{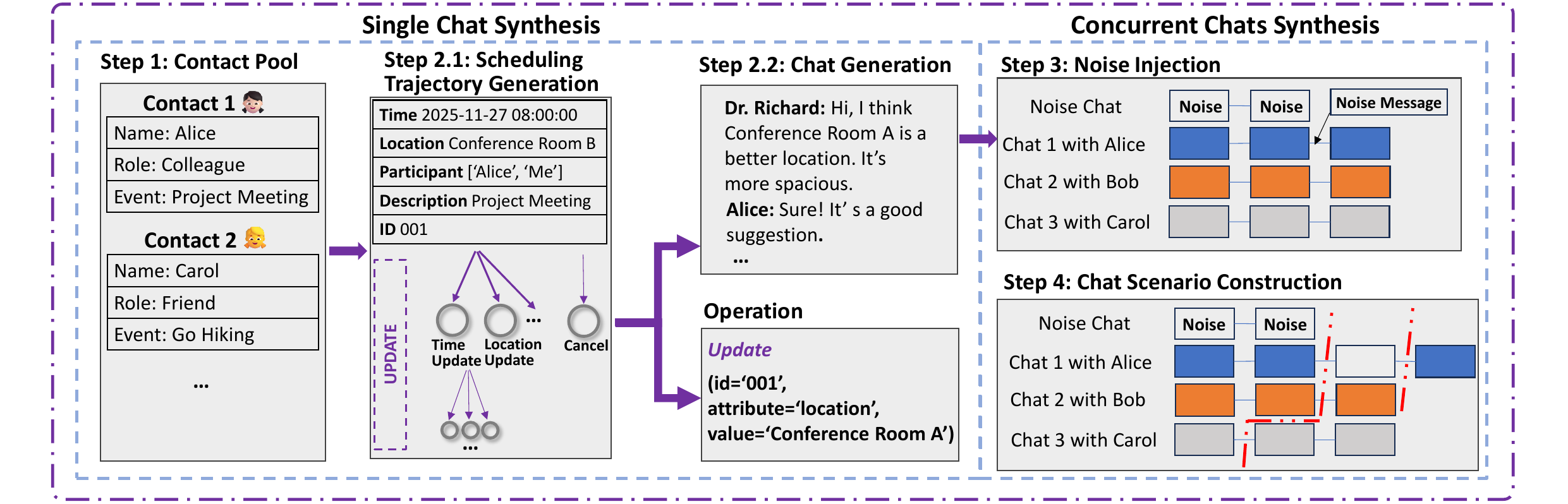}
  \caption{\textsc{ProEvent} construction pipeline. \textbf{Left:} Single chat synthesis. A contact is selected from the contact pool, and a scheduling trajectory is generated to update an attribute of the event. This trajectory then guides the chat generation, producing a single chat. \textbf{Right:} To simulate real-world concurrent chat threads, multiple chats are combined, noise is injected, and the chats are segmented into fixed time windows.}
  \label{fig:pipeline}
\end{figure*}

\subsection{Data Construction}

To better understand \textsc{ProEvent}, we first introduce its construction pipeline, which underpins the analysis of its diversity and quality.

\paragraph{Step one: Contact Pool Construction.}
To ensure consistency in the generated dialogues, we predefine a profile for each contact, including their name, role, and events to plan. For example, a colleague is more likely to schedule a project meeting, whereas a sports enthusiast may arrange a hiking.
To promote diversity and realism of these profiles, we draw inspiration from existing benchmarks~\cite{ye2022multiwoz,rastogi2020towards,quan2020risawoz,zhang2018personalizing} and define $56$ different profiles (Appendix~\ref{app:a}).

\paragraph{Step two: Chat Synthesis.}
At this stage, we synthesize chats involving selected contacts from the contact pool. To simulate the dynamic interactions of real--world conversations, we generate dialogues for negotiating and updating planned events with varying numbers of turns. We first create a scheduling trajectory that records the true state of all events at each time step. This trajectory then guides the chat generation, ensuring that each dialogue aligns with the underlying event timeline.

% \emph{2.1 Scheduling Trajectory Generation.}

To generate sequential scheduling trajectories that capture how events evolve over time, we start from an initialized event and iteratively modify its attributes, including time, location, and participants. To enhance the flexibility and diversity of the generated trajectories, we define three atomic operations to model these modifications~\cite{xu2024can,mathur2025sometimes}:
(1) \emph{Simple Modification}, which simulates negotiation adjustments (e.g., “I have another meeting at 10 A.M., so can we change the time to 5 P.M.?”);
(2) \emph{Timetable Linking}, which references existing or historical events in the timetable (e.g., “Let’s meet in the same conference room as last time”); and
(3) \emph{Detail Removal}, which represents tentative events with incomplete information (e.g., “Let’s meet tomorrow; I’ll provide the specific location later”).
The outcome of each operation can be either successful or unsuccessful. By randomly selecting an atomic operation and a target attribute at each turn, we generate scheduling trajectories with diversity and realistic variations.

% \emph{2.2 Chat Generation.}  
Based on the generated scheduling trajectory, we construct a corresponding chat skeleton that serves as a high-level blueprint for dialogue generation (Appendix~\ref{constru_case_app}). 
% For example, when the meeting location is updated to “Room A”, the skeleton may specify that a participant proposes this change. 
Finally, we use \texttt{GPT-OSS-120B} to generate dialogues that strictly adhere to the ground truths in the scheduling trajectory. 

\paragraph{Step three: Noise Injection.} 
Noise is an important factor in real-world proactive agent applications. To construct realistic test cases, we inject three types of noise into all generated chats: 
(1) \emph{Message-level noise}, messages irrelevant to the target event but interleaved with relevant messages within a chat; 
(2) \emph{Chat-level noise}, an entire chat unrelated to the target event; and 
(3) \emph{Event-level noise}, where additional determined or historical events in the timetable may distract the agent from operating on the correct event. 
In terms of content, we diversify noise to include off-topic messages, discussions about events unrelated to the user, and failed attempts to schedule events~\cite{li2025towards,higashinaka2021integrated}. This variety ensures the evaluation in a realistic chat environment.
Besides, to better simulate realistic conversational styles, we further inject slang expressions, discourse markers, and emoticons.
% ~\cite{gao2024dr3, li2025towards,higashinaka2021integrated}
\paragraph{Step Four: Chat Scenario Construction.}

To simulate concurrent chat threads and the periodic context collection setting in the real world, we merge multiple chats and segment them into fixed time windows.
Based on these time windows and the generated scheduling trajectories, we derive ground truth operations for each time step by comparing event states across successive windows. 
In this way, we construct scenarios which require agents to handle multiple chats and events simultaneously while obtaining the ground truth operation at each time step.

\begin{figure*}[t]
  \includegraphics[width=1\linewidth]{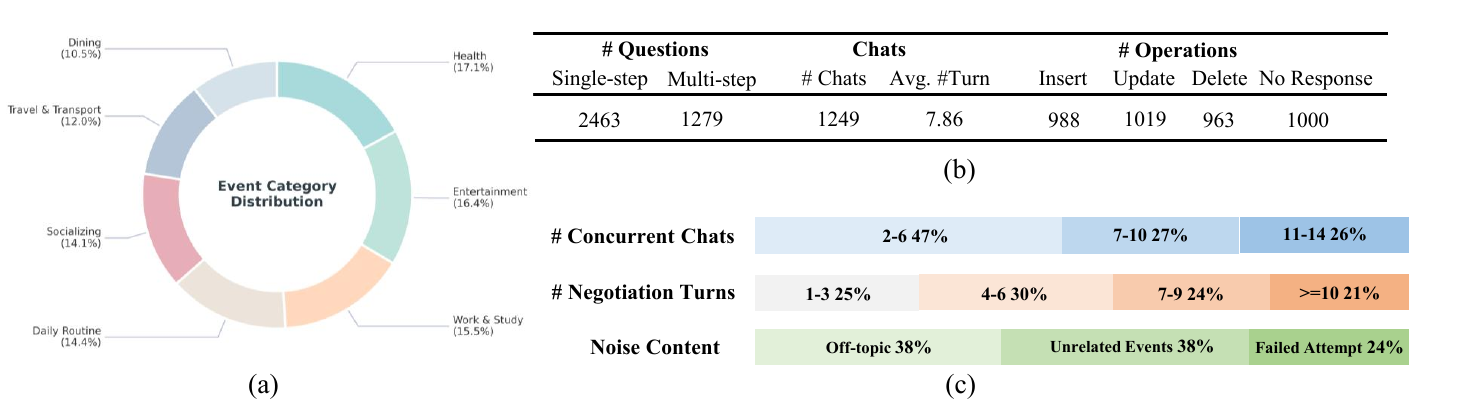}
  \caption{Dataset Statistics. (a) illustrates the diversity of events and their corresponding chat topics. (b) presents the benchmark statistics. (c) shows the distribution of negotiation turns, concurrent chats, and noise content, reflecting realistic chat characteristics.}
  \label{sta}
\end{figure*}

\subsection{Diversity}

\textsc{ProEvent} contains $2463$ single-step questions and $1279$ multi-step cases derived from $1249$ high-quality chats (Figure~\ref{sta}(b)).
By aggregating the event categories covered in existing task-oriented dialogue datasets~\cite{ye2022multiwoz,rastogi2020towards,quan2020risawoz}, \textsc{ProEvent} covers $7$ event categories and $56$ subcategories, providing broad coverage of real-world events.

% To further evaluate dataset diversity, we also measure the diversity of event scheduling patterns by analyzing the distribution of scheduling trajectories and measuring the diversity of linguistic expressions by manually analyzing a sampled subset. Comparing \textsc{ProEvent} with the widely used manually annotated task-oriented dialogue dataset MultiWOZ, \textsc{ProEvent} demonstrates higher diversity.
To further evaluate dataset diversity, we analyze \textbf{event scheduling patterns} through the distribution of scheduling trajectories and assess \textbf{linguistic variation} through manual analysis on a sampled subset of chats.
Compared with Multiwoz~\cite{ye2022multiwoz}, one of widely used task-oriented datasets, \textsc{ProEvent} demonstrates higher diversity across all evaluated dimensions (Appendix~\ref{app:a}).

In addition, to better reflect real-world chatting scenarios, \textsc{ProEvent} includes conversations with varying numbers of concurrent chats, different negotiation lengths, and diverse noise types.
The distribution is illustrated in Figure~\ref{sta}(c).

% \subsection{Diversity}

% \textsc{ProEvent} contains 2,049 single-step questions derived from 807 high-quality chats, 1,827 events and 622 timetables (Figure~\ref{sta} (b)). Regarding operation types, \emph{Delete} operations are substantially less frequent than \emph{Insert} and \emph{Update}. This is because a \emph{Delete} operation can occur at most once after an event is inserted, whereas insertions may occur without deletion, and multiple updates can be applied to the same event. To ensure this imbalance does not bias the results, we additionally evaluate models on a balanced subset, which yields consistent findings with the full dataset (Appendix~\ref{balanced}).

% Figure~\ref{sta} (a) demonstrates substantial diversity in event categories. In addition, to reflect real-world chat characteristics, \textsc{ProEvent} includes scenarios with varying numbers of concurrent chats, different numbers of negotiation turns, and diverse types of noise. The proportions of these categories are illustrated in Figure~\ref{sta} (c).

\subsection{Quality}

We verify the \textbf{correctness} of \textsc{ProEvent} through both automatic rule-based validation and human evaluation.
We first validate the structual consistency and semantic correctness via automatic validation.
For \textit{structural consistency}, we check whether all operations are executable in sequence and whether speakers are correctly aligned with participants.
For \textit{semantic correctness}, we validate whether scheduled event attributes are consistent with the corresponding chats and timetables.
Then, two annotators independently examine all cases that fail the automatic validation.
A case is retained only if both annotators judge it to be correct.
The annotators achieve $100\%$ agreement (Appendix~\ref{app:correctness}).

Besides, we further analyze the realism of the synthesized dialogues.
Using GPT-5.4 as a judge, we conduct pairwise comparisons between chats from \textsc{ProEvent} and dialogues from MultiWOZ~\cite{ye2022multiwoz} and Ubuntu Dialogue Corpus~\cite{lowe2015ubuntu}.
The results show that GPT-5.4 struggles to reliably distinguish synthesized chats from real dialogues (Appendix~\ref{app:realism}).

%% file: Body/Experiment.tex
\section{Experiment}

% \subsection{Setup}

We use \textsc{ProEvent} to evaluate LLM-based proactive agents across three categories: (1) open-source LLMs: including Qwen-3~\cite{yang2025qwen3}, Deepseek-V3.2~\cite{liu2025deepseek}, and Deepseek-R1~\cite{guo2025deepseek}; (2) proprietary LLMs: GPT-5.1; and (3) proactive-agent pipelines: Proactive, which encourages LLMs to respond more proactively, and ProCoT, which explicitly requires the model to reason about whether a response is necessary~\cite{deng2023prompting}.

Since we do not observe single-step ambiguity among operation types (e.g., an \emph{Update} may equal an \emph{Insert} combined with a \emph{Delete}), we just report precision and recall for single-step performance. For multi-step evaluation, an event is no longer tracked once an incorrect prediction occurs. Moreover, while time and participant fields can be exactly matched to determine correctness, we use LLMs as auxiliary judges to evaluate the correctness of open-ended location descriptions(Appendix~\ref{location}).

\begin{table*}[htbp]
    \centering
    \footnotesize
    \setlength{\tabcolsep}{2.5pt}
    \begin{tabular}{l cc cccccccccc}
    
    \toprule
    \multirow{3}{*}{\textbf{Models}} & \multicolumn{2}{c}{\multirow{2}{*}{\textbf{Timing}}} & \multicolumn{8}{c}{\textbf{Single-Step}} & \multicolumn{2}{c}{\textbf{Multi-Step}} \\
    
    % 关键：这里只保留第4-13列的cmidrule，不包含2-3列
    \cmidrule(lr){4-11} \cmidrule(lr){12-13}
    
    & \multicolumn{2}{c}{} & \multicolumn{2}{c}{Insert} & \multicolumn{2}{c}{Update} & \multicolumn{2}{c}{Delete} & \multicolumn{2}{c}{Overall} & \multicolumn{2}{c}{Success Rate} \\
    
    % 这里才是所有列的分隔线
    \cmidrule(lr){2-3} \cmidrule(lr){4-5} \cmidrule(lr){6-7} \cmidrule(lr){8-9} \cmidrule(lr){10-11} \cmidrule(lr){12-13}
    
    & FDR $\downarrow$ & MNR $\downarrow$ & R & P & R & P & R & P & R & P & Event & Table \\
    
    \midrule

    Qwen-3 (8B) & $48.1\%$ & $22.6\%$ & $41.8\%$ & $17.6\%$ & $29.8\%$ & $34.5\%$ & $11.5\%$ & $77.9\%$ & $27.5\%$ & $24.7\%$ & $30.4\%$ & $\underline{7.8\%}$ \\
    Qwen-3 (32B) & $\underline{96.6\%}$ & $21.7\%$ & $28.5\%$ & $\underline{9.3\%}$ & $12.2\%$ & $15.7\%$ & $19.0\%$ & $27.7\%$ & $19.8\%$ & $13.3\%$ & $22.3\%$ & $\underline{5.9\%}$ \\

    \midrule
    
    Qwen-3 (235B) & $77.5\%$ & $13.8\%$ & $42.1\%$ & $16.5\%$ & $32.4\%$ & $16.2\%$ & $37.5\%$ & $36.2\%$ & $37.3\%$ & $20.1\%$ & $38.7\%$ & $13.6\%$ \\
    
    DeepSeek-V3.2 & $\underline{96.5\%}$ & $21.9\%$ & $44.8\%$ & $20.6\%$ & $39.0\%$ & $19.3\%$ & $54.1\%$ & $38.4\%$ & $46.1\%$ & $24.7\%$ & $47.0\%$ & $18.5\%$ \\
    
    DeepSeek-R1 & $42.7\%$ & $5.4\%$ & $\textbf{61.5}\%$ & $30.4\%$ & $61.6\%$ & $52.3\%$ & $47.3\%$ & $97.1\%$ & $56.7\%$ & $46.3\%$ & $\textbf{57.7\%}$ & $\textbf{27.2\%}$ \\
    
    GPT-5.1 & $\textbf{24.0\%}$ & $14.5\%$ & $59.8\%$ & $\textbf{41.3\%}$ & $64.1\%$ & $49.6\%$ & $50.0\%$ & $77.1\%$ & $58.0\%$ & $51.5\%$ & $57.3\%$ & $26.7\%$ \\
    
    \midrule
    
    Proactive (Deepseek-V3.2) & $\underline{96.9\%}$ & $\textbf{1.4\%}$ & $46.2\%$ & $18.2\%$ & $37.2\%$ & $16.3\%$ & $\textbf{55.1\%}$ & $33.9\%$ & $46.1\%$ & $21.5\%$ & $47.6\%$ & $18.1\%$ \\
    
    ProCoT (Deepseek-V3.2) & $26.5\%$ & $8.5\%$ & $64.1\%$ & $38.3\%$ & $\textbf{71.8\%}$ & $\textbf{60.6\%}$ & $38.7\%$ & $\textbf{99.5\%}$ & $\textbf{58.2\%}$ & $\textbf{53.9\%}$ & $56.9\%$ & $25.9\%$ \\

    \midrule

    Human Performance & $0\%$ & $1.6\%$ & $95.7\%$ & $94.7\%$  & $97.8\%$ & $99.2\%$ & $96.9\%$ & $100\%$ & $96.9\%$ & $97.5\%$ & $94.4\%$ & $90.5\%$ \\
    
    \bottomrule
    \end{tabular}
    
    \caption{Evaluation results on \textsc{ProEvent}. $\downarrow$ indicates that lower values are better, while all other metrics are higher-is-better. We highlight \underline{problematic} results (FDR and MNR $>90\%$, while all other metrics are $<10\%$) and the \textbf{best-performing} results except humans. P and R denote Precision and Recall, respectively.}
    \label{tab:fixed_table}
\end{table*}

\subsection{Results Analysis}

% \paragraph{Overall performance.}
% GPT-5.1 achieves the best performance in almost all dimensions, with a False Detection Rate (FDR) of $38\%$ for response timing, an overall recall of $66.1\%$ for single-step response correctness (Table~\ref{tab:fixed_table}). Despite this advantage, the multi-step results indicate that GPT-5.1 can only correctly deal with about half of the events and provides trustworthy services in merely $30\%$ of cases. Similarly, DeepSeek-V3.2 achieves an Event Success Rate of only around $20\%$ and a Timetable Success Rate below $10\%$, whereas human annotators reach $94.4\%$ and $90.5\%$. The results underscore that \textbf{current LLMs remain far from delivering dependable proactive assistance } and that \textbf{\textsc{ProEvent} presents a challenging benchmark}.

\paragraph{Overall performance.}
Models with strong reasoning capabilities achieve substantially better performance on \textsc{ProEvent}.
In particular, ProCoT achieves the best overall single-step correctness ($58.2\%$ recall and $53.9\%$ precision), while DeepSeek-R1 obtains the highest Event and Timetable Success Rates ($57.7\%$ and $27.2\%$).
Despite these improvements, a substantial gap remains between single-step correctness and multi-step success.
Even GPT-5.1 can only correctly deal with about half of the events and provides trustworthy services in merely $26.7\%$ of cases.
The results underscore that \textbf{current LLMs remain far from delivering dependable proactive assistance} and \textbf{\textsc{ProEvent} presents a challenging benchmark}.

\paragraph{LLMs tend to overreact.} 
Across nearly all evaluated models, the FDR is significantly higher than the Missed Need Rate (MNR), suggesting a systemic bias toward over-responsiveness. 
For instance, DeepSeek-V3.2 exhibits an FDR more than $70\%$ higher than its MNR ($96.5\%$ vs. $21.9\%$). 
These results indicate that LLMs frequently trigger proactive actions when no intervention is required, yet they are relatively more capable of avoiding omissions when assistance is actually needed. 
% This ``hallucination of needs'' can impose unnecessary cognitive and interaction burdens on users. 
Furthermore, we observe that the ProCoT pipeline---which explicitly prompts the model to evaluate the necessity of an action before execution---reduces DeepSeek-V3.2’s FDR by $70\%$ (from $93.4\%$ to $40.8\%$). This suggests that deliberative reasoning about response necessity mitigates LLMs’ inherent tendency to over-anticipate user needs.

% \paragraph{Stronger models are more prudent in deleting events.}
% GPT-5.1, DeepSeek-R1, and the ProCoT strategy achieve the highest overall response correctness, along with the lowest FDR and MNR. Notably, these models also exhibit the lowest recall for \textit{Delete} operations (26.7\%, 22.8\%, and 28.2\%, respectively). 
% In contrast, although the weaker DeepSeek-V3.2 attains a substantially higher recall for \emph{Delete} operations, its precision is only 18.0\%, compared with 90.0\% for GPT-5.1. This discrepancy indicates that models with stronger response timing judgment and higher overall correctness tend to adopt a more conservative strategy when deciding to delete events, avoiding deletions unless there is clear evidence. 
% A similar bias has also been observed in prior studies, where LLMs tend to avoid commonsense reasoning~\cite{li2024mvp,fu2025mme} and show a strong preference in binary Yes/No questions,
% % A similar cautious behavior is also observed when models engage in proactive commonsense reasoning. 
% In \textsc{ProEvent}, such prudence is particularly relevant to the scenarios where users quit from a planned activity, or when cancellation cues are subtle (Section~\ref{discussion}).

\paragraph{Reasoning improves recognition of implicit event cancellations.}
Identifying when to \textit{Delete} an event is particularly challenging because cancellations are often expressed implicitly rather than through explicit user instructions.
Although DeepSeek-V3.2 and the Proactive strategy achieve relatively high recall for delete operations ($54.1\%$ and $55.1\%$), their low precision ($38.4\%$, $33.9\%$) indicates that they frequently remove events without sufficient evidence.
In contrast, GPT-5.1, DeepSeek-R1, and ProCoT achieve substantially higher precision ($77.1\%$, $97.1\%$, $99.5\%$) while maintaining competitive recall, suggesting that stronger reasoning capability helps models better distinguish genuine cancellations from temporary or ambiguous schedule changes.

\begin{figure}[ht] % 建议使用 ht 或 !t 增加灵活性
  \centering % 加上这行，确保图片在栏内水平居中
  \includegraphics[width=\linewidth]{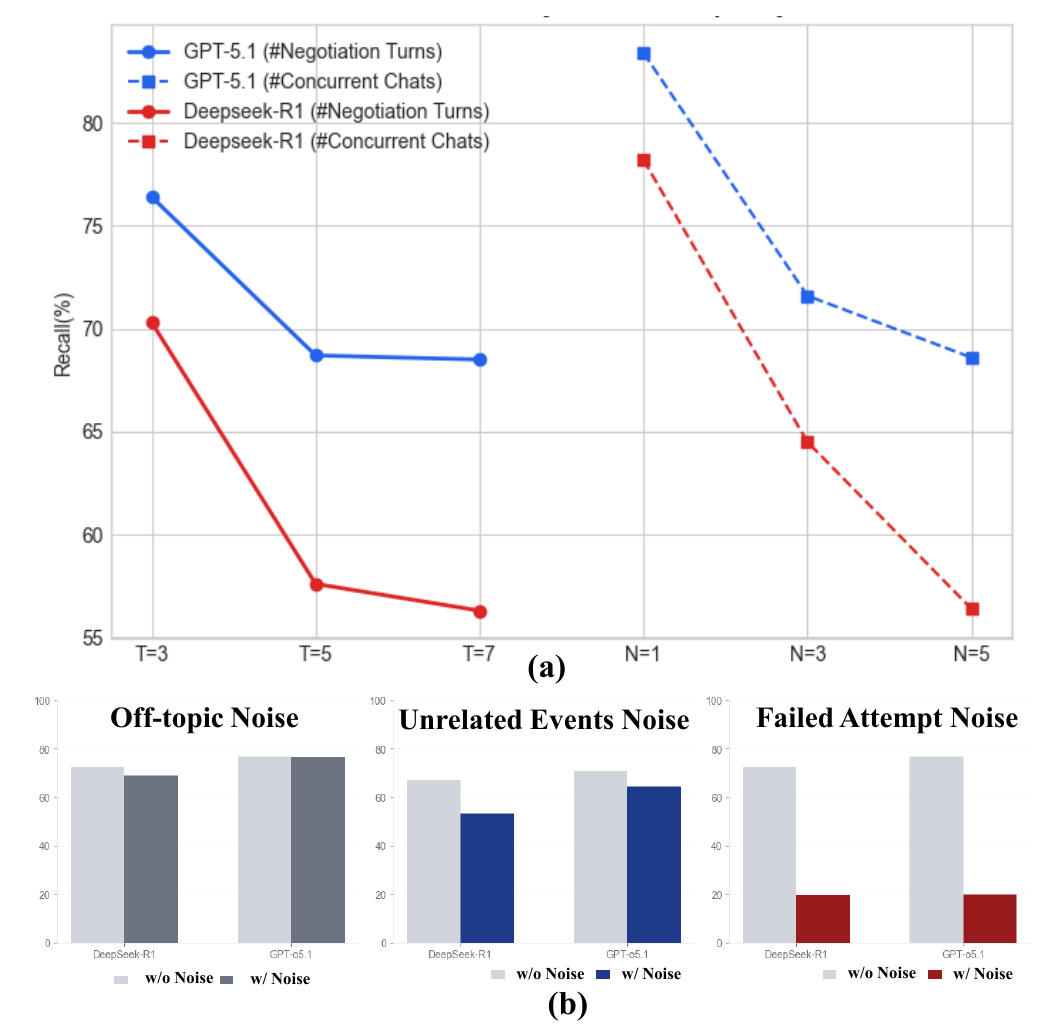} % 稍微放大一点
  \caption{Effects of dynamic interactions ((a), left), concurrent chats ((a), right), and noise (b) on the performance of Deepseek-R1 and GPT-5.1. Here, $T$ denotes the number of negotiation turns, and $N$ denotes the number of concurrent chats. The experiments are conducted on $300$ sampled cases.}
  \label{discussion_robustness} % 建议 label 起名更有辨识度
\end{figure}

\paragraph{Real-world chat complexities pose significant challenges for LLMs.}
Since \textsc{ProEvent} incorporates dynamic user interactions, concurrent chat threads, and diverse forms of noise, we examine how these factors affect model performance. Our results show that recall consistently declines across all evaluated models as the number of negotiation turns increases and the volume of concurrent threads grows (Figure~\ref{discussion_robustness} (a)).
With respect to noise, we find that \textbf{the semantic content of noise affects LLMs more than its positioning}. While injecting off-topic noise at varying levels does not significantly degrade performance (Appendix~\ref{real-world}), replacing such noise with discussions of events unrelated to the user or failed planning attempts leads to substantial drops in recall and precision, respectively (Figure~\ref{discussion_robustness}(b)).
These findings suggest that while LLMs can understand the semantics of chats, they still struggle to robustly infer underlying human intentions in complex, realistic chat settings.

\subsection{Discussion}
\label{discussion}

In this section, we present qualitative analysis observations and discuss LLMs' deficiencies as proactive agents.
\begin{figure}[ht] % 建议使用 ht 或 !t 增加灵活性
  \centering % 加上这行，确保图片在栏内水平居中
  \includegraphics[width=\linewidth]{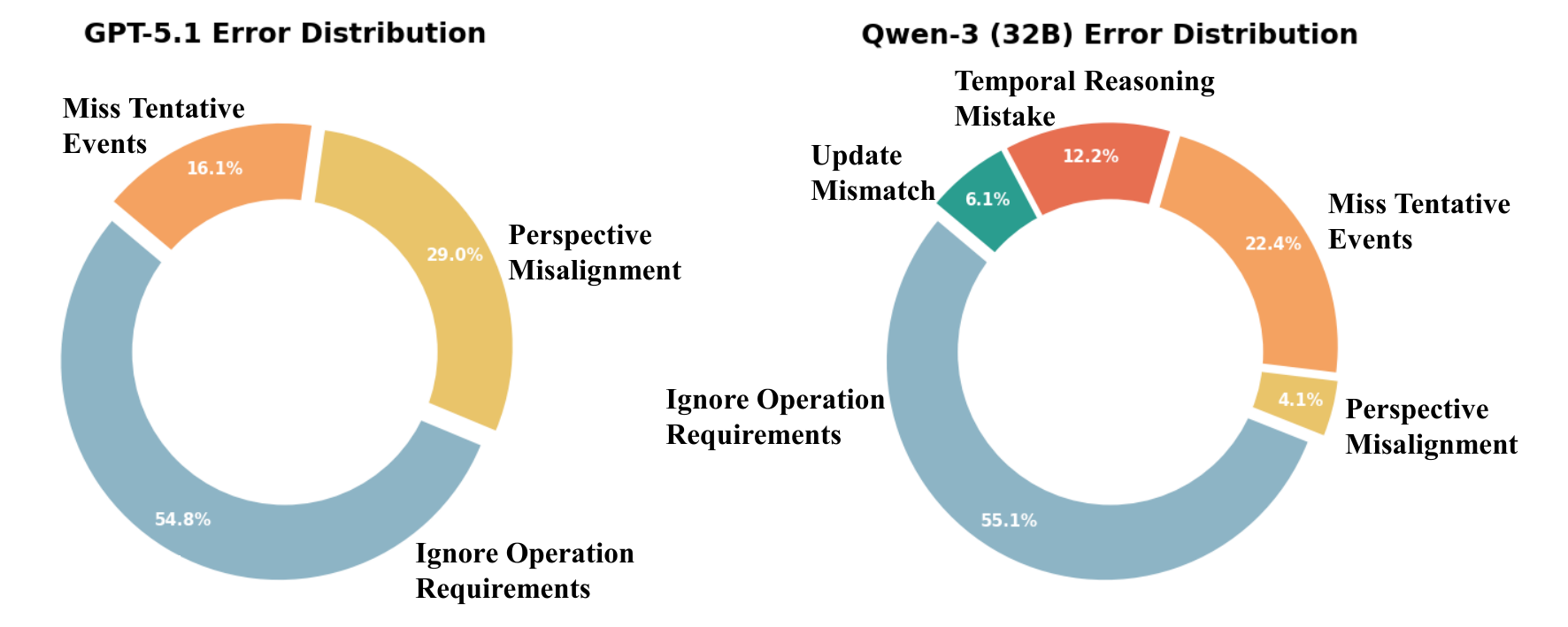} % 稍微放大一点
  \caption{Error distributions of GPT-5.1 and Qwen-3.}
  \label{fig:error_dist} % 建议 label 起名更有辨识度
\end{figure}

% \begin{figure*}[t]
%   \includegraphics[width=\linewidth]{Figure/Discussion_fig1.pdf}
%   \caption{}
%   \label{exp-dis}
% \end{figure*}

\begin{figure*}[t]
    \centering
  \includegraphics[width=0.88\linewidth]{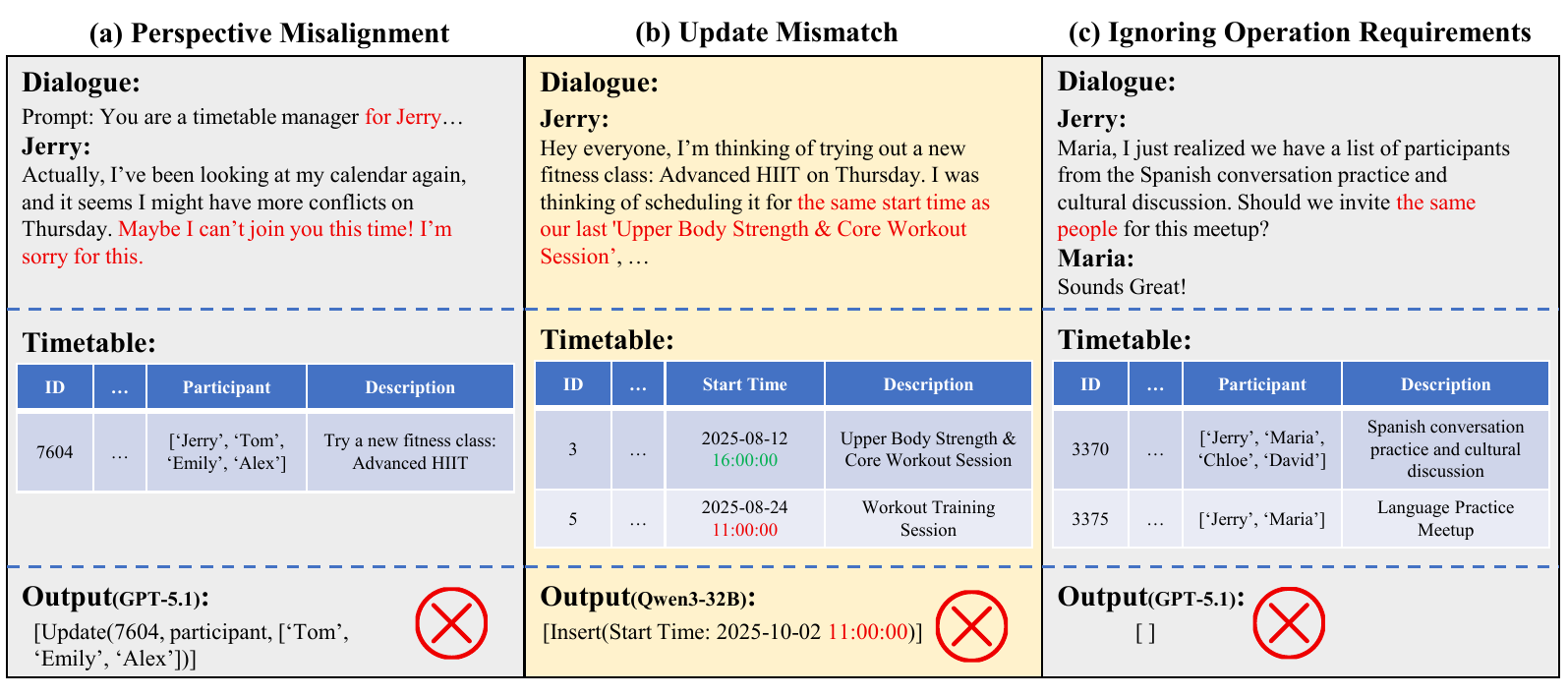}
  \caption{Error cases. LLMs make mistakes when reasoning from the user’s first-person perspective, updating one of multiple events in the timetable, or missing required operations implicitly implied by the chats. Cases (a) and (c) are collected from GPT-5.1, while case (b) is collected from Qwen-3-32B. Additional cases involving temporal reasoning errors and missing tentative events are provided in Appendix~\ref{case_app}.}
  \label{casestudy}
\end{figure*}

\paragraph{GPT-5.1 is prone to overlooking events implicitly expressed in narrative styles.} 
Despite its state-of-the-art reasoning capabilities, GPT-5.1 exhibits a notable failure rate when task instructions are embedded in natural, narrative dialogue. For instance, in Figure~\ref{casestudy} (c), the model fails to trigger a participant update when a user provides the implicit instruction to ``invite the same people''. Given the model's high performance in structurally explicit tasks, we hypothesize that the cause is not a lack of reasoning capacity, but rather a perceptual bias where the model ignores operational requirements in the absence of explicit structural cues. 

To verify this assumption, we conducted an ablation study on a curated set of these error cases by introducing two interventions: (1) substituting narrative instructions with direct expressions (e.g., specific names), and (2) appending an update prompt to the dialogue (see Appendix~\ref{ablation}). As shown in Table~\ref{discussion_tab_1}, both interventions markedly improve GPT-5.1’s performance. Notably, the update prompt increases Recall by $32.0\%$, an effect not observed in other models. These results confirm that while GPT-5.1 possesses the underlying reasoning capabilities, its performance as a proactive agent is highly sensitive to the explicitness of information within the context.

\begin{table}
    \centering
    \small
    \begin{tabular}{l cc}
         \hline
         
           & \textbf{\small Recall} & \textbf{\small Precision} \\

           \hline
           GPT-o5.1 & $56.0\%$ & $77.8\%$ \\
           GPT-o5.1 + direct expression & $76.0\%$ & $90.5\%$ \\
           GPT-o5.1 + update prompt & $88.0\%$ & $88\%$ \\
           \mydeltax{$\Delta$\hfill} & \mydelta{$\textbf{32\%}$} & \mydelta{$\textbf{10.2\%}$} \\ 

           \hline
           Qwen-3 (32B) & $44.0\%$ & $55.0\%$ \\
           Qwen-3 (32B) + direct expression & $29.2\%$ & $70.0\%$ \\
           Qwen-3 (32B) + update prompt & $44.0\%$ & $91.7\%$ \\
           \mydeltax{$\Delta$\hfill} & $0\%$ & \mydelta{$\textbf{36.7\%}$} \\ 

           % \hline
           % LLaVA-1.5 (7B) & $51.74$ & $68.89$ \\
           % LLaVA-1.5 (7B) + VC & $53.48$ & $69.26$ \\
           % \mydeltax{$\Delta$\hfill} & \mydelta{$+1.74$} & \mydelta{$+0.37$} \\ 

           % \hline
           % GPT-4V & $39.57$ & $66.30$ \\
           % GPT-4V+VC & $43.91$ & $64.81$ \\
           % \mydeltax{$\Delta$\hfill} & \mydelta{$+4.34$} & \textcolor{red}{\mydelta{$-1.49$}} \\ 

           % \hline
           % GPT-4o & $56.09$ & $74.44$ \\
           % GPT-4o+VC & $58.70$ & $75.19$ \\
           % \mydeltax{$\Delta$\hfill} & \mydelta{$+2.61$} & \mydelta{$+0.75$} \\ 

           \hline
           
    \end{tabular}
    \caption{Effect of making implicit narratives explicit and adding an update prompt to the chats. $\Delta$ denotes the performance gain, and we highlight \textbf{improved results}.}
    \label{discussion_tab_1}
    \vskip -0.1in
\end{table}

\paragraph{Crucial Reasoning Abilities for Proactive Agents.}
By comparing the error distributions of GPT-5.1 and Qwen-3-32B (Figure~\ref{fig:error_dist}), we identify two error categories that are present in Qwen-3 but absent in GPT-5.1: temporal reasoning mistakes and update mismatch. This contrast reveals two critical reasoning abilities that distinguish stronger models from weaker ones: temporal reasoning and memory-enhanced reasoning.
% Temporal reasoning is fundamental, as time constitutes a core attribute of an event.
In Qwen-3, temporal reasoning errors primarily arise from incorrect transformations between dates and weekdays, leading to erroneous scheduling decisions. This suggests that models lacking robust temporal reasoning may benefit from auxiliary tools, such as calendar or calculator utilities, to support precise temporal computation~\cite{li2023api,parisi2022talm}.
Furthermore, as illustrated in Figure~\ref{casestudy} (b), update mismatch refers to cases where the agent modifies an incorrect timetable entry or retrieves information from an unrelated historical event. This error reflects a deficiency in memory-enhanced reasoning, which requires the agent to accurately retrieve and align relevant information across multiple events.
% Together, these observations suggest that effective proactive agents must go beyond surface-level intent understanding and incorporate robust temporal reasoning and memory-enhanced reasoning to maintain consistent and correct behavior in dynamic, concurrency environments.

\paragraph{LLMs lack a first-person user perspective when making decisions.}
To provide effective assistance in dynamic and complex environments, proactive agents must consistently reason from the user’s first-person perspective. However, our analysis reveals a systematic failure of LLMs to make decisions from the user-centric viewpoint. As illustrated in Figure~\ref{casestudy} (a), when the user (Jerry) states that he will no longer attend an activity, the correct action is to perform a \emph{Delete} operation that removes the event from Jerry’s personal timetable. Instead, the model frequently performs an \emph{Update} operation that merely removes “Jerry” from the participant list while retaining the event itself.
This indicates that LLMs reason about events from a third-person perspective rather than adopting the user--centered viewpoint~\cite{hou2024egosocialarena,cheng2024egothink}. This limitation poses a challenge for deploying LLMs as reliable proactive agents.

% \paragraph{LLMs Struggle to deal with uncertainty.}
% The omission of tentative events constitutes a significant failure mode for both GPT-o5.1 and Qwen-32B. As illustrated in Figure~\ref{casestudy}(c), users often make soft commitments—for example, scheduling an activity on a particular day while leaving the exact time to be determined. Although such tentative entries are critical for maintaining an accurate and proactive timetable, LLM-based agents frequently fail to record them.
% This limitation arises from two primary factors. First, LLMs struggle to represent and reason about uncertainty, tending to either overcommit to fully specified events or ignore partially specified ones altogether~\cite{xiong2023can}. Second, models often misinterpret user intent in tentative planning scenarios, treating provisional statements as non-actionable rather than as signals that warrant lightweight or incomplete event creation.

%% file: Body/Conclusion.tex
\section{Conclusion}

We introduce \textsc{ProEvent}, the first event-centric benchmark designed to evaluate proactive agents’ ability to track users’ upcoming events from instant messaging chats. Through a comprehensive evaluation of eight LLMs, we diagnose their systematic biases in both response timing and content. Our analyses further reveal fundamental deficiencies of current LLMs as proactive agents, particularly in real-world application scenarios.
We hope that \textsc{ProEvent} will facilitate future research on enhancing LLMs’ capabilities for proactive event tracking in real-world settings.

%% file: Appendix/Appendix_A.tex
\appendix
\onecolumn

\section{Diversity of \textsc{ProEvent}}
\label{app:a}

We evaluate the diversity of \textsc{ProEvent} from three aspects:
(1) event category diversity,
(2) event scheduling pattern diversity, and
(3) linguistic diversity.
To provide a reference point, we compare \textsc{ProEvent} with MultiWOZ, a widely used task-oriented dialogue dataset that also contains event-centric scheduling conversations.

\paragraph{Event Category Diversity.}
We measure event category diversity by counting the number of covered event categories.

\paragraph{Event Scheduling Pattern Diversity.}
Both MultiWOZ and \textsc{ProEvent} can be represented as sequences of scheduling actions.
We therefore treat each unique trajectory as a scheduling pattern and measure the number of distinct scheduling patterns.

\paragraph{Linguistic Diversity.}
To evaluate linguistic diversity, we consider the linguistic variations used in three common scheduling intents:
\textit{propose}, \textit{modify}, and \textit{cancel}.
For each intent, we randomly sample 50 messages from \textsc{ProEvent} and MultiWOZ and manually count the number of distinct linguistic expressions (e.g., ``I'm looking to'' for propose or ``Can we switch'' for modify).

As shown in Table~\ref{app:div}, \textsc{ProEvent} demonstrates broader event coverage, more diverse scheduling patterns, and richer linguistic variations than MultiWOZ.

\begin{table}[t]
\centering
% \footnotesize
\setlength{\tabcolsep}{4pt}

\begin{tabular}{l cc ccc}

\toprule

\multirow{2}{*}{\textbf{Dataset}} &
\multirow{2}{*}{\textbf{Event Categories}} &
\multirow{2}{*}{\textbf{Scheduling Patterns}} &
\multicolumn{3}{c}{\textbf{Linguistic Variations (50 Samples)}} \\

\cmidrule(lr){4-6}

& & &
\textbf{Propose} &
\textbf{Modify} &
\textbf{Cancel} \\

\midrule

Multiwoz & 4 & 55 & 13 & 17 & 7 \\

\textsc{ProEvent} & 7 & 275 & 18 & 27 & 14 \\

\bottomrule

\end{tabular}

\caption{Diversity comparison between Multiwoz and \textsc{ProEvent}.}
\label{app:div}

\end{table}

\section{\textsc{ProEvent} Correctness Annotation}
\label{app:correctness}

\begin{figure*}[t]
  \includegraphics[width=1\linewidth]{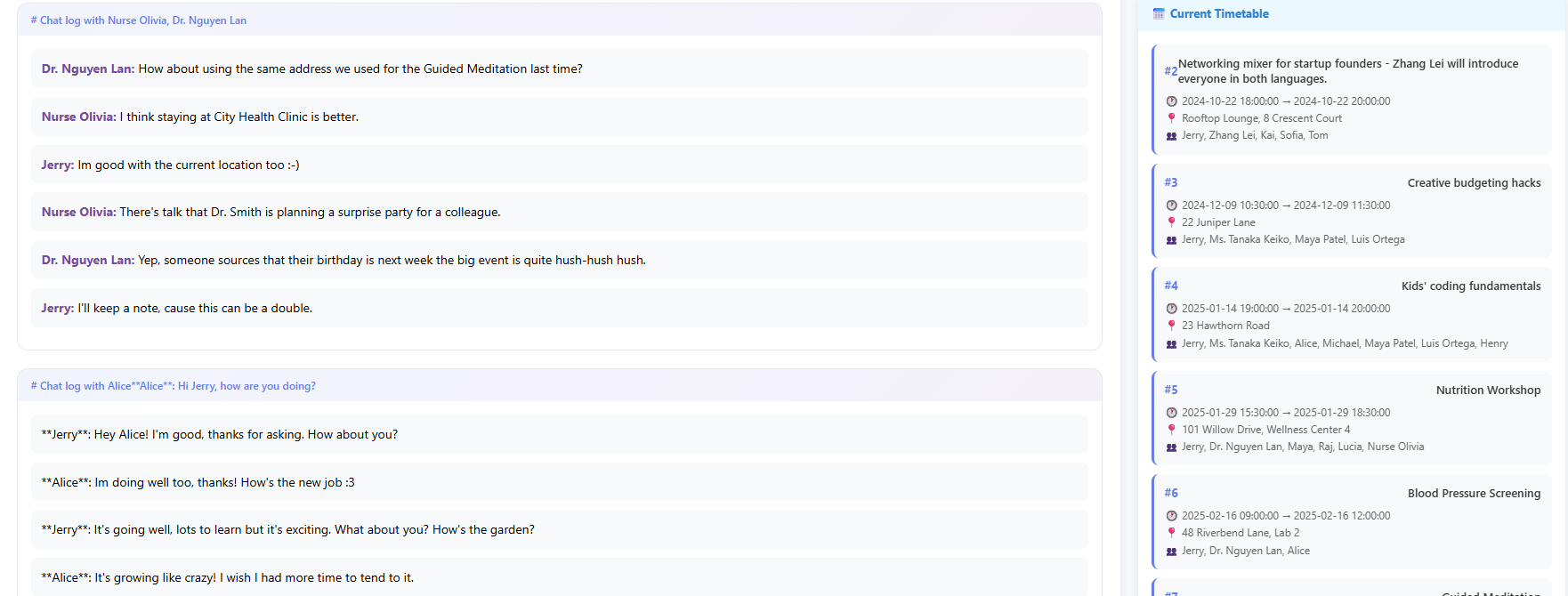}
  \caption{Pairwise realism results between ProEvent and realistic datasets.}
  \label{annotate_website}
\end{figure*}

Automatic rule-based validation mainly relies on string matching, which may lead to overly strict judgments in semantically equivalent cases.
For example, when the ground-truth location is Conference Room, Third Floor'', the corresponding dialogue may express it as the conference room on the third floor''.
Similarly, when the ground truth is Anna's kitchen'', the dialogue may contain expressions such as Anna: In my kitchen''.
Therefore, we additionally conduct human annotation to verify semantic correctness.
The annotation interface is shown in Figure~\ref{annotate_website}.

\section{\textsc{ProEvent} Realism Evaluation}
\label{app:realism}

We evaluate the realism of \textsc{ProEvent} via pairwise comparisons against real-world dialogue datasets, including MultiWOZ and the Urban Dialogue Corpus.

For each comparison, we randomly sample a message from \textsc{ProEvent} and a message from the real-world dataset, and ask GPT-5.4 to identify which one is more likely to be from the real world.
To reduce order bias, we repeat the evaluation with reversed input orders, and only treat a prediction as valid when both orders yield consistent results.

To control for superficial cues, we further match samples by dialogue length and punctuation statistics during sampling.

\begin{figure*}[t]
  \includegraphics[width=1\linewidth]{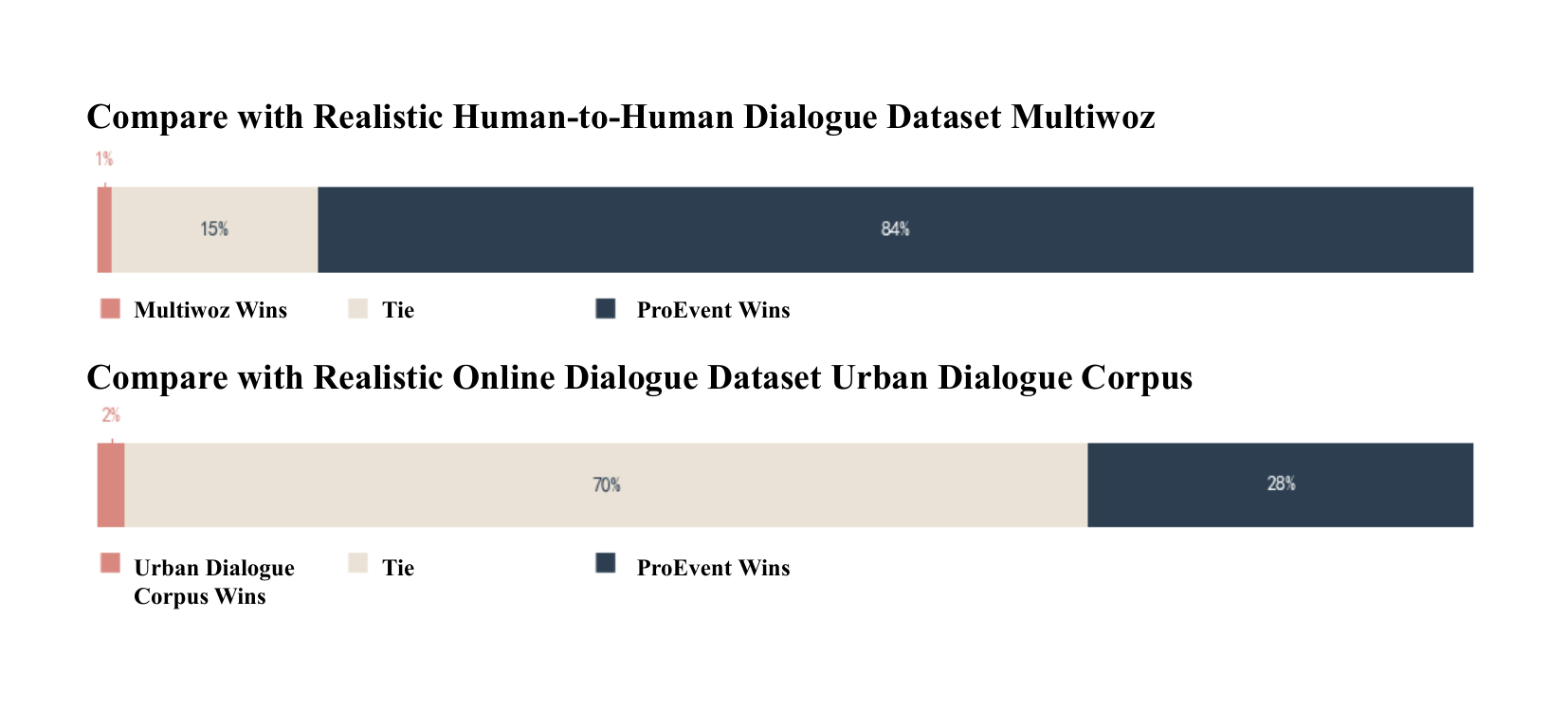}
  \caption{Pairwise realism results between ProEvent and realistic datasets.}
  \label{app_rea1}
\end{figure*}

As shown in Figure~\ref{app_rea1}, GPT-5.4 struggles to consistently distinguish \textsc{ProEvent} from both MultiWOZ and the Urban Dialogue Corpus, suggesting that \textsc{ProEvent} exhibits realistic human conversational characteristics across both task-oriented and online dialogue settings.

Besides, we further analyze whether synthesized dialogues lead to substantially different model behaviors compared with human-written dialogues.
Specifically, we sample $50$ test cases from \textsc{ProEvent} and manually rewrite the dialogues while preserving the original intent of each message.
Results show that only $6$ cases produce different model outputs between the rewritten and synthesized dialogues, whereas repeated runs on the synthesized dataset alone already result in $8$ inconsistent cases.
This suggests that the impact of synthesized dialogue styles is smaller than the intrinsic randomness of LLM inference.

\section{A case for the explicit expression and the update prompt}
\label{ablation}

\begin{figure*}[]
  \includegraphics[width=1\linewidth]{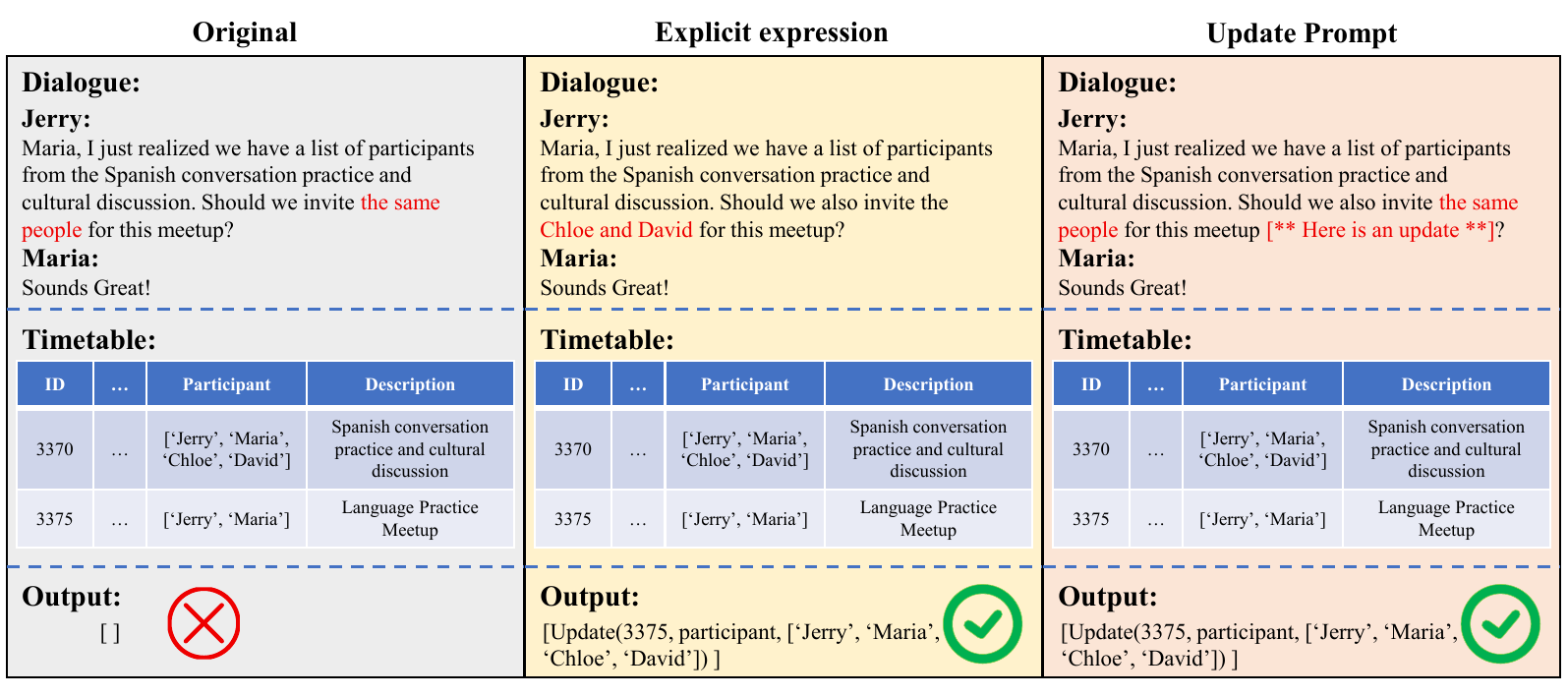}
  \caption{A case for demonstrating the effect of explicit expression and the update prompt.}
  \label{app-f-1}
\end{figure*}

Figure~\ref{app-f-1} demonstrates a case for explicit expression and the update prompt. The case shows that after explicitly mentioning the names of participants or adding a prompt to remind LLMs to update both correct the original wrong answers.

\label{appendix_b}

\section{LLMs' robustness on real-world chat complexities}
\label{real-world}

The results in Table~\ref{tab:negotiation_results_extended} demonstrate that almost all models' performance degrades when the number of negotiation turns and the number of concurrent chats grow. As for noise, the effect of off-topic noise is negligible, while the discussions on an event unrelated to the user and a failed attempt to plan an event lead to a significantly lower recall and precision, respectively.

\begin{table}[h]
    \setlength{\tabcolsep}{1.5pt} % 稍微缩小列间距以适应增加的列
    \centering
    \small
    \begin{tabular}{l ccc ccc cccccccc}
        \toprule
        \multirow{3}{*}{\textbf{Models}} & \multicolumn{3}{c}{\textbf{Negotiation Turn}} & \multicolumn{3}{c}{\textbf{Concurrent Chat}} & \multicolumn{8}{c}{\textbf{Noise}} \\
        \cmidrule(lr){2-4} \cmidrule(lr){5-7} \cmidrule(lr){8-15}
        & \multirow{2}{*}{T=3} & \multirow{2}{*}{T=5} & \multirow{2}{*}{T=7} & \multirow{2}{*}{N=1} & \multirow{2}{*}{N=3} & \multirow{2}{*}{N=5} & \multicolumn{2}{c}{w/o} & \multicolumn{2}{c}{w/ OT} & \multicolumn{2}{c}{w/ UE} & \multicolumn{2}{c}{w/ FA} \\
        \cmidrule(lr){8-9} \cmidrule(lr){10-11} \cmidrule(lr){12-13} \cmidrule(lr){14-15}
        & & & & & & & R & P & R & P & R & P & R & P \\
        \midrule
        DeepSeek-V3.2 & $46.5\%$ & $34.2\%$ & $27.6\%$ & $50.0\%$ & $44.4\%$ & $30.0\%$ & $53.2\%$ & $58.3\%$ & $58.9\%$ & $64.7\%$ & $53.2\%$ & $46.7\%$ & $54.4\%$ & $19.9\%$ \\
        DeepSeek-R1 & $70.3\%$ & $57.6\%$ & $56.3\%$ & $78.2\%$ & $64.5\%$ & $56.4\%$ & $67.1\%$ & $72.6\%$ & $60.8\%$ & $69.1\%$ & $53.2\%$ & $54.5\%$ & $50.6\%$ & $19.5\%$ \\
        GPT-5.1 & $\textbf{76.4\%}$ & $\textbf{68.7\%}$ & $\textbf{68.5\%}$ & $\textbf{83.4\%}$ & $\textbf{71.6\%}$ & $\textbf{68.6\%}$ & $\textbf{70.9\%}$ & $\textbf{76.7\%}$ & $\textbf{69.6\%}$ & $\textbf{76.4\%}$ & $\textbf{64.6\%}$ & $\textbf{73.9\%}$ & $\textbf{60.8\%}$ & $19.8\%$ \\
        \midrule
        % Direct (Deepseek-V3.2) & $46.5\%$ & $34.2\%$ & $27.6\%$ & $50.0\%$ & $44.4\%$ & $30.0\%$ & $53.2\%$ & $58.3\%$ & $58.9\%$ & $64.7\%$ & $53.2\%$ & $46.7\%$ & $54.4\%$ & $19.9\%$ \\
        Proactive (Deepseek-V3.2) & $45.9\%$ & $35.3\%$ & $28.5\%$ & $48.3\%$ & $43.2\%$ & $32.8\%$ & $51.9\%$ & $55.4\%$ & $61.4\%$ & $68.5\%$ & $49.4\%$ & $34.8\%$ & $49.4\%$ & $17.3\%$ \\
        ProCoT (Deepseek-V3.2) & $63.8\%$ & $55.1\%$ & $51.8\%$ & $53.6\%$ & $46.6\%$ & $44.6\%$ & $63.3\%$ & $72.5\%$ & $65.2\%$ & $75.8\%$ & $59.5\%$ & $65.3\%$ & $63.3\%$ & $\textbf{23.7\%}$ \\
        \bottomrule
    \end{tabular}
    \caption{LLMs' performance with different negotiation turn numbers, concurrent chat numbers, and diversifying noise. R refers to Recall, while P refers to Precision. OT refers to the off-topic messages. UE means the discussions about an event unrelated to the user. FA means a failed attempt to plan an event. We highlight the models' \textbf{best performance}.}
    \label{tab:negotiation_results_extended}
\end{table}

\section{Cases on temporal reasoning mistakes and tentative events}
\label{case_app}

\begin{figure*}[t]
  \includegraphics[width=1\linewidth]{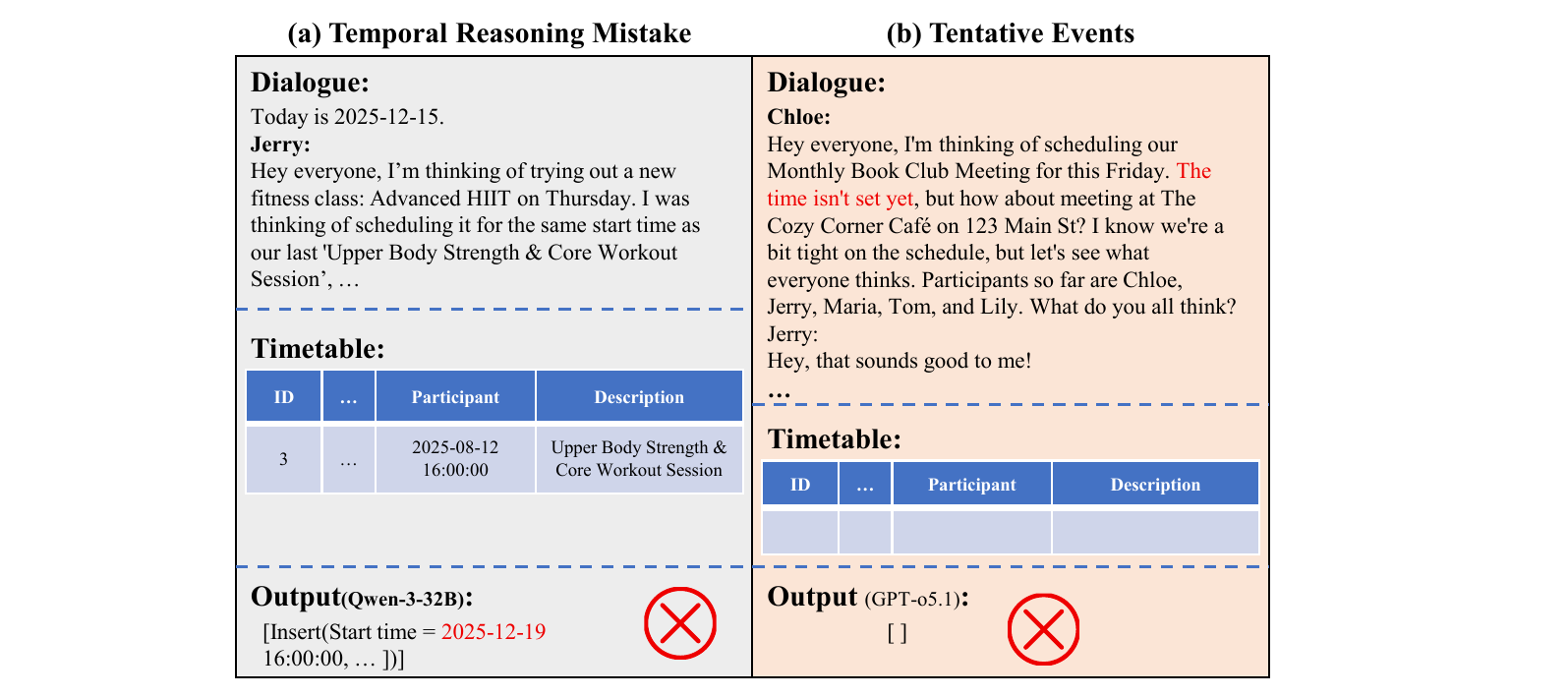}
  \caption{The case for temporal reasoning mistakes and tentative events.}
  \label{case_app}
\end{figure*}

Figure~\ref{case_app} (a) demonstrates a case for temporal reasoning mistakes, when Qwen3-32B fails to transform the weekday into a date. As for the tentative events (Figure~\ref{case_app} (b)), LLMs fail to insert an event when the date is set and the time is to be determined. This reflects that LLMs may lack an ability to handle uncertainty.

% The omission of tentative events constitutes a significant failure mode for both GPT-o5.1 and Qwen-32B. As illustrated in Figure~\ref{casestudy}(c), users often make soft commitments—for example, scheduling an activity on a particular day while leaving the exact time to be determined. Although such tentative entries are critical for maintaining an accurate and proactive timetable, LLM-based agents frequently fail to record them.
% This limitation arises from two primary factors. First, LLMs struggle to represent and reason about uncertainty, tending to either overcommit to fully specified events or ignore partially specified ones altogether~\cite{xiong2023can}. Second, models often misinterpret user intent in tentative planning scenarios, treating provisional statements as non-actionable rather than as signals that warrant lightweight or incomplete event creation.

\section{Judging the Correctness of Location with LLMs}
\label{location}

The description of a location can be diverse and linguistically varied even when referring to the same physical entity. For example, a meeting might be described as occurring in ``Room 305,'' ``the third-floor conference room,'' or ``305 Meeting Hall.'' Such variations make exact string matching an unreliable metric for evaluation. Hence, we use GPT-5.1 with a 4-shot setting to handle the location's open-ended nature and evaluate the semantic equivalence between the predicted location and the ground truth. 

The prompt is designed to provide the model with the task instruction and chat context. The specific prompt structure is detailed as follows. Besides, we compare the judgment consistency between GPT-5.1 and humans on 100 randomly sampled cases; the results show that they are $\textbf{100\%}$ consistent, validating the reliability of using an LLM as a proxy evaluator for this task.

\begin{verbatim}
You are a meticulous assistant tasked with evaluating the correctness of a proactive
agent's location extraction.

Input:
- Ground Truth Location: [ground truth location]
- Predicted Location: [predicted location]
- Chat Context: [chat context related to the location]

Task:
Determine whether the predicted location refers to the same specific place as the
ground-truth location. Consider abbreviations, synonyms, hierarchical descriptions,
and cases where the location is left implicit.

Output Format:
Return **CORRECT** if the two locations refer to the same place; otherwise, return
**INCORRECT**.
\end{verbatim}

\section{A case of our chat synthesis pipeline}
\label{constru_case_app}

\begin{figure}[]
  \includegraphics[width=\linewidth]{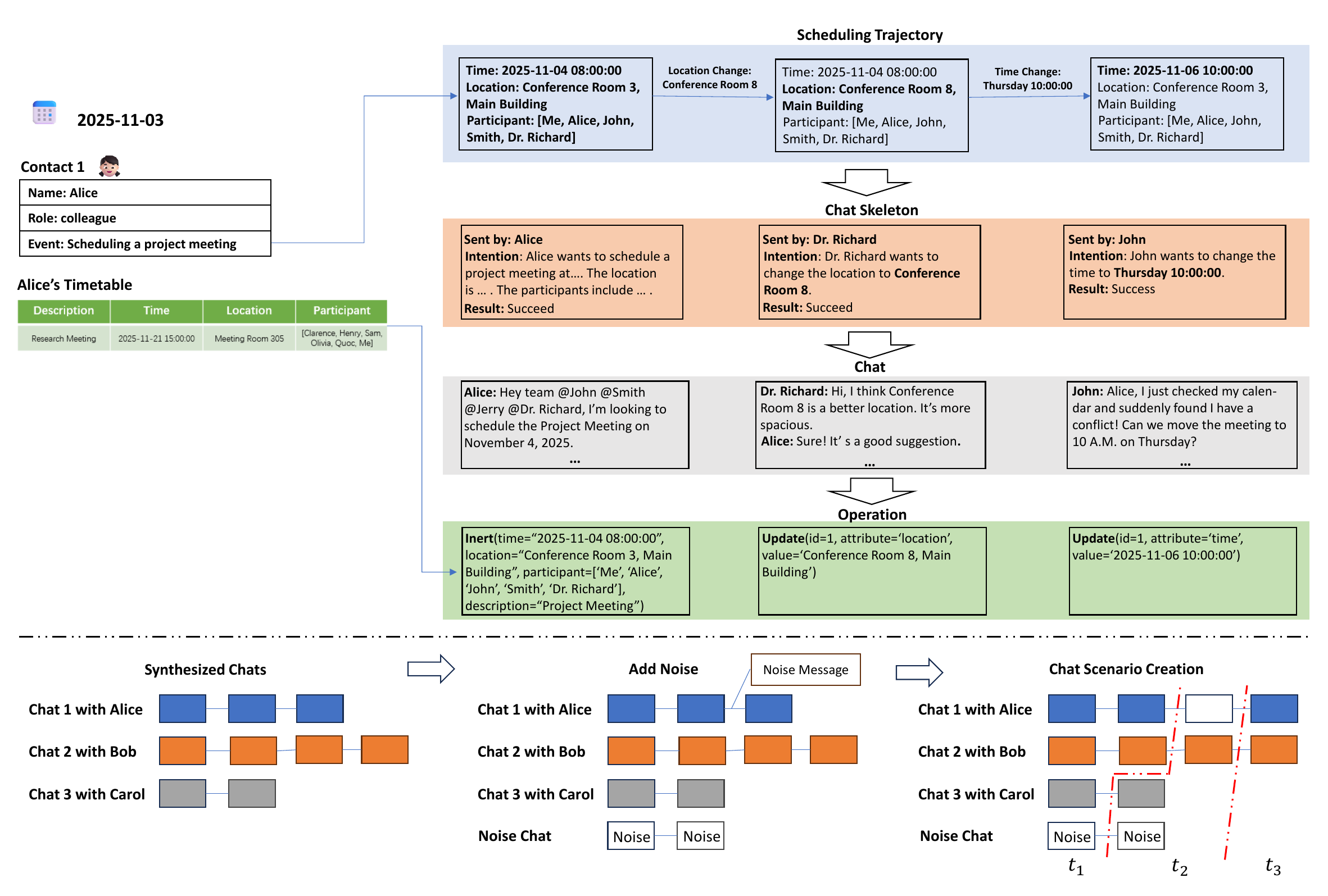}
  \caption{A specific case of our chat synthesis pipeline.}
  \label{constru_case}
\end{figure}

Figure~\ref{constru_case} illustrates a concrete example of our chat synthesis pipeline. We first select a contact from the contact pool and generate scheduling trajectories to simulate the event planning process. Based on these trajectories, we construct a chat skeleton by specifying the event update proposer and organizing updates in a narrative form. Guided by the skeleton, we generate the chat and derive ground-truth operations by tracking changes in the scheduling trajectory. Finally, the lower part of Figure~\ref{constru_case} shows how multiple chats are combined to form a concurrent chat scenario.

\section{The prompts for evaluating LLMs}

We use the following prompts to evaluate LLMs and provide four examples (\emph{Insert}, \emph{Update}, \emph{Delete} and no response) for further clarifying the definition of each operation. The prompt can also be found in the questions of our dataset.

\begin{verbatim}
You are a scheduling manager for Jerry. Given a dialogue and the current timetable,
identify the required schedule operations.

Rules:
- INSERT a new event when a previously unscheduled event is confirmed.
- UPDATE an existing event when its start time, end time, location, or participants change.
- DELETE an event when it is explicitly cancelled.
- Output [] if no operation is required.

The current timetable contains confirmed and historical events and serves as contextual memory.
If any information is unspecified, leave the corresponding field empty.
If the date is fixed but the time is unknown, output the date only.
Return only the operation results without any explanation.

Operation formats:
1. INSERT: (INSERT, {"start_time": "%Y-%m-%d %H:%M:%S",
                     "end_time": "%Y-%m-%d %H:%M:%S",
                     "location": str,
                     "participant": list,
                     "description": str})

2. UPDATE: (UPDATE, {"id": int,
                     "attribute": str,
                     "value": str})

3. DELETE: (DELETE, {"id": int})
\end{verbatim}